\documentclass[preprint,12pt]{elsarticle}

\usepackage{amssymb,algorithm,algorithmic,graphicx,color}

\newtheorem{definition}{Definition}[section]
\newtheorem{lemma}{Lemma}[section]
\newtheorem{prop}[lemma]{Proposition}
\newtheorem{cor}[lemma]{Corollary}

\newtheorem{example}[lemma]{Example}
\newtheorem{alg}[lemma]{Algorithm}
\newtheorem{thm}[lemma]{Theorem}

\newcommand{\scst}{\scriptscriptstyle}

\begin{document}
\begin{frontmatter}

\title{A Tool for  Integer Homology Computation: $\lambda$-AT-Model}
\author{R. Gonzalez-Diaz, }\author{M.J. Jimenez, }\author{B. Medrano}\author{, P. Real}

\address{Applied Math Department, University of Sevilla,\\
Campus Reina Merdedes, CP: 41012, Sevilla, Spain\\
$\{$rogodi,majiro,belenmg,real$\}$@us.es\\
http://alojamientos.us.es/gtocoma}

\begin{abstract}

In this paper, we formalize the notion of $\lambda$-AT-model (where $\lambda$ is a non-null integer)
 for a given chain complex, which allows the
computation of homological information in the integer domain
avoiding using the Smith Normal Form of the boundary
matrices. We present an  algorithm for computing such a
model, obtaining Betti numbers, the prime numbers $p$ involved in
the invariant factors of the torsion subgroup of homology, the
amount of invariant factors that are a power of $p$ and a set of
representative cycles of  generators of homology mod $p$, for
each $p$. Moreover, we establish  the minimum valid
$\lambda$ for such a construction, what cuts down the
computational costs related to the torsion subgroup. The tools
described here are useful to determine topological information of nD
structured objects such as simplicial, cubical or simploidal
complexes and are applicable to extract such an information from
digital pictures.

\end{abstract}

\end{frontmatter}

\section{Introduction}
Important questions in Vision and Image Processing involve an
accurate topological analysis. In some scientific areas, such as
Medical Imaging and Remote Sensing, there is a growing need to
visualize, analyze and manipulate data of high complexity and
dimensionality. We are concerned about providing algorithms to
solve topological problems in any dimension.
 For this aim, efficient methods
from combinatorial or algebraic topology to compute topological
features are needed. Due to the fact that we will work with
n-dimensional combinatorial objects (mainly, $n=3,4...$),
topological features are not limited to Betti numbers, Euler
characteristic, connectivity, number of holes or cavities, but
also include other advanced characteristics such as (co)homology
in the integer domain, cohomology ring, cohomology operations
or homotopy groups, which can help to discriminate topologically
non-equivalent objects. Algebraic Topology and Homological Algebra
fields provide the necessary tools to capture all this
topological stuff (see \cite{Munkres,Mac95,F87}).

The classical algorithm for computing
 homology groups (in the integer domain) uses the Smith Normal Form (SNF) of the
 boundary
matrices of a chain complex. Explicit examples can be given for which the computation of 
SNF has a worst-case computational complexity which grows exponentially in both
space and time \cite{GH97}. Many algorithms have been devised
 to improve this complexity bound
 \cite{Iliopoulos,Storjohann,Dumas,Peltier}.
 
In the AT-model theory \cite{GR03,GR05}, starting from a
simplicial complex $K$, a chain homotopy equivalence is generated
from the chain complex associated to $K$, ${\cal C}(K)$, to its
homology. In this case, the  complexity of the associated algorithm  is
${\cal O}(m^3)$, where $m$ is the number of generators of the
complex, but the computation is carried out over a field.
In this
paper, we extend the notion of AT-model for integer homology
computation. More concretely, we define
the homological algebra notion of $\lambda$-AT-model and describe
an algorithm for computing a structure of this kind from any chain
complex, with coefficients in the integer domain (${\bf Z}$). 
Starting from this
information, it is possible, in particular, to obtain Betti
numbers, the prime numbers $p$ involved in the invariant factors
(corresponding to the torsion subgroup of the homology), the
amount of invariant factors that are a power of $p$ and a set of
representative cycles of  generators of homology mod $p$, without using SNF.
We  present an algorithm for computing such a $\lambda$-AT-model for a given chain complex and extracting all this homological information, performed in ${\cal O}(m^3 \psi(\lambda))$ in the worst case, $\psi$ being the Euler function.

In the following section, we recall classical definitions from
Algebraic Topology. In Section 3, previous tools for computing
topological information such as AM-models  (which make use of 
SNF), and AT-models (for computing homological information over
a field) are given. In Section 4, we define the notion of
$\lambda$-AT-model, study its properties, give an algorithm for
computing it and study its complexity. Finally, we describe how
to extract homological information in the integer domain from
$\lambda$-AT-models. The last section is devoted to conclusions
and future work.

\section{Background}
This section introduces the basic concepts and definitions needed
throughout the paper, taking as main reference Munkres' book
\cite{Munkres}.

A \textit{chain complex} ${\cal C}$ is a sequence
$\cdots\,\stackrel{d_{q+1}}{\rightarrow} {\cal
C}_{q}\stackrel{d_q}{\rightarrow} {\cal
C}_{q-1}\stackrel{d_{q-1}}{\rightarrow}\, \cdots
\,\stackrel{d_1}{\rightarrow} {\cal
C}_0\stackrel{d_0}{\rightarrow}0$ of free abelian groups ${\cal C}_q$
and homomorphisms $d_q:{\cal C}_{q}\to {\cal C}_{q-1}$, such that,
for all $q$,  $d_q d_{q+1}=0$.
An element $a\in {\cal C}_q$ is a $q$-{\em chain} of ${\cal C}$; we say that $q$ is the dimension of $a$, and write $q=dim(a)$.
The set $d=\{d_q\}$ is
called the \textit{differential} of ${\cal C}$.
If there is no confusion, we can write $a\in {\cal C}$ if $a\in {\cal C}_q$ for some $q$. Similarly,
we can omit the subindex $q$ in $d_q(a)$.

The chain complex ${\cal C}$ is  {\em finite} if there exists an integer $n>0$ such that
${\cal C}_q=0$ for $q>n$ and each abelian group ${\cal C}_q$ is finitely generated. In this case, if
${\cal C}_n\neq 0$, we say that dim of ${\cal C}$ is $n$.

Since our goal is the computation of homological information of
finite objects,  all chain complexes considered here are finite. In this case, ${\cal C}$ can be encoded as a pair
$(C,d)$, where: (1) $C=\{C_q\}$, and $C_q$ is  a set of
generators of ${\cal C}_q$; (2) $d=\{d_q\}$ and for each $q$,
$d_q$ is the differential of ${\cal C}$ in dim $q$  with
respect to the bases $C_q$ and $C_{q-1}$. 
Sometimes, we  write ${\cal C}=(C,d)$ if the chain complex ${\cal C}$ is encoded as the pair $(C,d)$.
If $b\in {\cal C}_q$ and $a\in C_q$, then $\langle b,a \rangle$ denotes
the coefficient of the generator $a$ in the $q$-chain $b$ when $b$ is
expressed as a linear combination of elements of $C_q$.

\begin{example}\label{ex}
Shapes are classically modeled with cellular subdivisions.
Several combinatorial structures may represent such a subdivision.
Simplicial complexes have proved to be a useful tool to model a
geometric object. Roughly speaking, they are collections of
simplices (vertices, edges, triangles, tetrahedra, ...) that fit
together in a natural way to form the object. A $q$-simplex
$\sigma$ is a convex hull of a set of $q$ affinely independent
points. If the set is $\{v_0,v_1,\dots,v_q\}$ then the simplex is
denoted by $v_0v_1\cdots v_q$. The boundary operator over the
$q$-simplex $\sigma$ is defined by $\partial_q(v_0v_1\cdots
v_q)=\sum_{i=0}^q(-1)^i v_0v_1\cdots\hat{v_i}\cdots v_q$ where
$\hat{v_i}$ means that the vertex $v_i$ is omitted. For every
simplicial complex $K$, one can define a chain complex ${\cal
C}(K)$ (canonically associated to it) such that the $q$-chains
are linear combinations of $q$-simplices. By linearity, the
boundary operator can be extended to chains to define the
differential $\partial$ of ${\cal C}(K)$.
 The homology of $K$ is then defined as the homology of
${\cal C}(K)$.
Another way of combinatorial representation of a geometric structure (which could arise naturally, for example,
 from tomography, numerical computations
and graphics), is by means of cubical grids, which subdivide the
space into cubes with vertices in an integer lattice. This
approach, that can be generalized to an arbitrary dimension,
leads to a cubical complex. The homology of a given cubical
complex is the homology of the cubical chain complex associated
to it \cite{Massey}. Finally, simploidal sets \cite{dahmen}
include simplicial complexes and cubical complexes as particular
cases. They can be used for
 representing `hybrid' grids coming from finite element
methods. In \cite{Peltier2}, a free chain complex is associated to a simploidal set and the homology of the simploidal set
is defined as the homology of the associated chain complex.
\end{example}

\begin{figure}[t!]
\begin{center}\includegraphics[width=8cm]{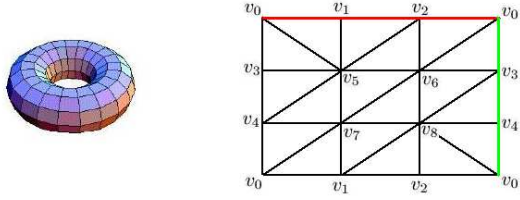}\end{center}
\caption{The Torus and a triangulation $T$ of it.}
\label{1}\end{figure}

\begin{figure}[t!]
\begin{center}\includegraphics[width=8cm]{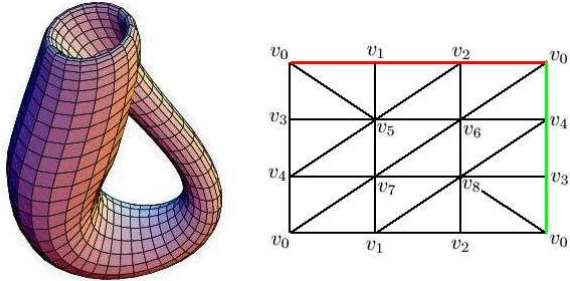}\end{center}
\caption{The Klein bottle and a triangulation $S$ of it.}
\label{2}\end{figure}

Given a chain complex ${\cal C}$ with differential $d$, a $q$--chain $a\in {\cal C}_q$ is called a
$q$--{\em cycle} if $d_q  (a)=0$.
If $a=  d_{q+1} (b)$ for some $b\in {\cal C}_{q+1}$ then $a$
 is called a $q$--{\em boundary}.
 Denote the groups of $q$--cycles
 and $q$--boundaries by $Z_q$ and $B_q$ respectively.
We say that two $q$-cycles $a$ and $b$ are {\em homologous} if there exists
a $(q+1)$-chain $c$ such that $a=b+d_{q+1}(c)$.
 Define the integer {\em $q$th  homology group} to be the quotient group
 $Z_q/B_q$, denoted by $H_q({\cal C};{\bf Z})$.

For each $q$, the integer $q$th homology group $H_q({\cal C};{\bf Z})$ is a finitely generated abelian group.
Moreover,
$H_q({\cal C};{\bf Z})$
is isomorphic to $F_q({\cal C};{\bf Z})\oplus T_q({\cal C};{\bf Z})$ where
$$F_q({\cal C};{\bf Z})={\bf Z}\oplus \cdots\oplus {\bf Z}\;\;\mbox{ and }\;\;T_q({\cal C};{\bf Z})=({\bf Z}/\alpha_{(q,1)})\oplus\cdots
\oplus({\bf Z}/\alpha_{(q,s)})$$ are the {\em free subgroup} and the {\em torsion subgroup} of
$H_q({\cal C};{\bf Z})$, respectively.
The rank of $F_q({\cal C};{\bf Z})$, denoted by $\beta_q$,  is called the $q$th {\em Betti number} of ${\cal C}$.
Intuitively,
 $\beta_0$ is the number of connected components, $\beta_1$ is the number of independent holes and
 $\beta_2$ is the number of cavities.
Each $\alpha_{(q,i)}$ is a power of a prime, $\alpha_{(q,i)}=p_i^{t_{(q,p_i)}}$.
They are called the {\em invariant factors} of $ H_q({\cal C};{\bf Z})$.
The numbers $\beta_q$ and  $\alpha_{(q,i)}$ are uniquely determined by $H_q({\cal C};{\bf Z})$
(up to a rearrangement). Therefore, this representation is in some sense a `canonical form' for $H_q({\cal C};{\bf Z})$.
For all $q$, there exists a finite number of elements of $ H_q({\cal C};{\bf Z})$
 from which we can deduce all the $H_q({\cal C};{\bf Z})$ elements. Those elements are called {\em homology generators} of
 dim $q$.
 We say that $a$ is a {\em representative $q$--cycle} of a homology generator $\alpha$ of the
 free subgroup of $H_q({\cal C};{\bf Z})$ if
 $\alpha=a+B_q$ and for each $\mu \in {\bf Z}$, $\mu \neq 0$, $\mu a \notin Im \, d$.
 We say that $a$ is a {\em representative $q$--cycle} of a homology generator $\alpha$ of the
 torsion subgroup of $H_q({\cal C};{\bf Z})$ if
 $\alpha=a+B_q$ and there exists $\mu \in {\bf Z}$, $\mu \neq 0$, such that $\mu a \in Im \, d$.
 We denote $\alpha=[a]$.

\begin{prop}\label{betti} \cite[p. 332]{Munkres} The $q$th homology group of ${\cal C}$ with coefficients in ${\bf
Z}/p$ for $p$ being a prime,  denoted by $H_q({\cal C};{\bf Z}/p)$, is
a vector space. Its rank, denoted by $\beta_{(q,p)}$, depends on
the prime $p$ and it is related to the number of invariant
factors of $H_q({\cal C};{\bf Z})$ that are a power of $p$,
$T_{(q,p)}$, via the Universal Coefficient Theorem for Homology,
which implies that, for each
prime $p$,
$$T_{(q,p)}=\beta_{(q,p)}-\beta_{q}-T_{(q-1,p)}\,\mbox{ for $q>0$}
\;\;\mbox{ and }\;\;T_{(0,p)}=\beta_{(0,p)}-\beta_{0}.$$ 
\end{prop}

\begin{example}\label{toroklein} Consider a triangulation $T$ of the torus and a triangulation $S$ of the Klein bottle (see Fig. \ref{1}).
Let ${\cal C}(T)$ and ${\cal C}(S)$ be the chain complexes canonically associated to $T$ and $S$, respectively.

The homology groups of the torus 
 are:
$$H_0({\cal C}(T);{\bf Z})\simeq {\bf Z}, \;H_1({\cal C}(T);{\bf Z})\simeq {\bf Z}\oplus {\bf Z},
\; H_2({\cal C}(T);{\bf Z})\simeq{\bf Z}.$$ The Betti numbers for the
torus are: $\beta^{\scst T}_0=1$, $\beta^{\scst T}_1=2$ and $\beta^{\scst T}_2=1$.
Representative cycles of homology generators of dim $1$ are shown in red and
green in Fig. \ref{1}, respectively. The $2$-chain $\tau$ which is the sum (up to signs) of all the triangles of
$T$ satisfies that $\partial_2(\tau)=0$, so $\tau$ is a
representative cycle of a homology generator of dim $2$.

In the case of the Klein bottle, its homology
groups are: $$H_0({\cal C}(S);{\bf Z})\simeq {\bf Z}, \;H_1({\cal C}(S);{\bf Z})\simeq {\bf
Z}\oplus {\bf Z}/2, \; H_2({\cal C}(S);{\bf Z})\simeq 0.$$ The Betti numbers of the Klein bottle are
$\beta^{\scst S}_0=1$, $\beta^{\scst S}_1=1$ and $\beta^{\scst S}_2=0$. A representative cycle
of a homology generator of the free subgroup  of
$H_1({\cal C}(S);{\bf Z})$ is shown in red in Fig.\ref{2}. The
invariant factor of $H_1({\cal C}(S);{\bf Z})$ is  $2$. The
 representative cycle $\alpha$ of a homology generator of
the torsion subgroup of $H_1({\cal C}(S);{\bf Z})$, which is
shown in green  in Figure \ref{2}, satisfies that $\partial_2(\gamma)=2 \alpha$ where 
 $\gamma$ is the sum (up to signs) of all the triangles of
$S$.

Working with coefficients in a field, it is not possible, in general, to get all the information
corresponding to the torsion subgroup of the homology. For
example, the homology groups of $S$ with coefficients in ${\bf
Z}/3$ are $$H_0({\cal C}(S);{\bf Z}/3)={\bf Z}/3, \;
H_1({\cal C}(S);{\bf Z}/3)={\bf Z}/3, \; H_2({\cal C}(S);{\bf Z}/3)=0$$ since
$\partial_2(\gamma)=2 \alpha$, so $\partial_2(2\gamma)= \alpha$.
Notice that the torsion part is lost since the
prime $p=3$ is not involved in any invariant factor of $H({\cal C}(S),{\bf Z})$.
 The homology groups of the Klein bottle with
coefficients in ${\bf Q}$ are
 $$H_0({\cal C}(S);{\bf Q})={\bf Q}, \; H_1({\cal C}(S);{\bf Q})={\bf Q}, \; H_2({\cal C}(S);{\bf Q})=0$$
 since $\partial_2(\frac{1}{2} \gamma)=\alpha$. Finally, the homology groups of the Klein bottle with
coefficients in ${\bf Z}/2$ are $$H_0({\cal C}(S);{\bf Z}/2)={\bf Z}/2, \;
H_1({\cal C}(S);{\bf Z}/2)={\bf Z}/2\oplus {\bf Z}/2, \; H_2({\cal C}(S);{\bf
Z}/2)={\bf Z}/2$$ since $\partial_2(\gamma)=2 \alpha =0$. 
 \end{example}

Let ${\cal C}$ and
${\cal C'}$ be two chain
complexes. Denote the differential of ${\cal C}$ and ${\cal C}'$ by $d$ and $d'$, respectively.
 A \textit{chain map} $f: {\cal C} \rightarrow {\cal
C'}$ is a family of homomorphisms $\{f_q: C_q \rightarrow C'_q\}$ such
that $d'_q f_q = f_{q-1} d_q$ for all $q\geq 0$.
If no confusion can arise, we omit the subindex and write $f(a)$ instead of  $f_q(a)$ for any $q$ and $a\in{\cal C}_q$.
A chain map $f: {\cal C} \rightarrow {\cal
C'}$ induces a homomorphism $[f]: H({\cal C};{\bf Z}) \rightarrow
H({\cal C'};{\bf Z})$ where $[f]([a])=[f(a)]$ for every $[a] \in H({\cal
C};{\bf Z})$.
If $f,g : {\cal C} \rightarrow {\cal C'}$ are chain maps,
then \textit{a chain homotopy} $\phi:{\cal C}\to {\cal C'}$
of $f$ to $g$ is a family of
homomorphisms $\{\phi_q : C_q \rightarrow C'_{q+1}\}$ such that $f_q-g_q=d'_{q+1}\phi_q + \phi_{q-1}
d_q $.
A \textit{chain contraction} of a chain complex ${\cal C}$,
 to
another chain complex ${\cal C'}$ with differentials $d$ and $d'$, respectively, is a set
$(f,g,\phi)$ such that: $f : {\cal C} \rightarrow {\cal C'}$ and
$g: {\cal C'} \rightarrow {\cal C}$ are chain maps; $fg$ is the
identity map of ${\cal C'}$ (denoted by $id_{\scst \cal C'}$)
and $\phi: {\cal C} \rightarrow {\cal
C}$ is a chain homotopy of the identity map of ${\cal C}$ to
$gf$, that is, $\phi d + d \phi = id_{\scst \cal C} - gf$.
Important properties of chain contractions are: (1)
$g$ is injective and $f$ is surjective since $fg=id_{\scst \cal C'}$ and, in particular,
  ${\cal C'}$ has fewer or the same number of generators than ${\cal C}$;
  (2) $[f][g]=id_{\scst H({\cal C}';{\bf Z})}$ and $[g][f]=id_{\scst H({\cal C};{\bf Z})}$ and, therefore,
   the homology groups of ${\cal C}$ and ${\cal C'}$ are isomorphic \cite{Massey}.

\begin{figure}[t!]
\begin{center}\includegraphics[width=8cm]{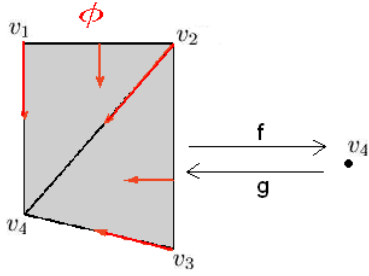}\end{center}
\caption{The chain contraction $(f,g,\phi)$ of ${\cal C}(K)$ to ${\cal C}(L)$.}
\label{3}\end{figure}

\begin{example} Let $K$ be the simplicial complex given by the following set of simplices (see Fig. \ref{3} on the left):
 $$\{v_1,v_2,v_3,v_4,v_1v_2,v_2v_3,v_3v_4,v_2v_4,v_1v_4,v_1v_2v_4,v_2v_3v_4\}$$
 and let $L$ be the simplicial complex given by the vertex $v_4$.

Let $f : {\cal C}(K) \rightarrow {\cal C}(L)$ and  $g : {\cal
C}(L) \rightarrow {\cal C}(K)$ be the chain maps given by
$f(v_i)=v_4$ for $i=1,2,3,4$ and $f$ is null over the other
simplices; $g(v_4)=v_4$.
Let $\phi  : {\cal C}(K) \rightarrow {\cal C}(K)$ be the chain
homotopy of $id_{\scst {\cal C}(K)}$ to $gf$ given by:
$\phi(v_1)=v_1v_4$, $\phi(v_2)=v_2v_4$, $\phi(v_3)=v_3v_4$, $\phi(v_1v_2)=v_1v_2v_4$, $\phi(v_2v_3)=v_2v_3v_4$ and
 $\phi$ is null over the other simplices.
The set $(f,g,\phi)$ is a chain contraction of ${\cal C}(K)$ to
${\cal C}(L)$ and hence, ${\cal C}(K)$ and ${\cal C}(L)$ have
isomorphic homology groups. Fig. \ref{3} illustrates such a chain contraction. Observe that the action
of the map $\phi$ `reduces' the simplicial complex to its
homology.
\end{example}

\section{AM-model and AT-model}
We recall here the concepts of AM-model and AT-model. Both tools
 represent the prior work that lays the foundations of $\lambda$-AT-model. In fact, the latter
 attempts to take the qualities of each of the former, computing homological information in the integer domain avoinding 
 the computation of SNF.

An {\em algebraic minimal model} (AM-model)  \cite{GR01} is a tool for computing topological information of chain complexes,
tool that is applicable to extract topological
  information from digital pictures. In particular, it provides integer (co)homology generators,
  representative (co)cycles of these generators as well as the cohomological invariant $HB1$ (derived from the rank of the cohomology ring) \cite{GR01,GR06,GR08}.
More concretely, an AM-model  for a chain complex ${\cal C}=(C,d)$  is
a set $((C,d), (M,d') f,g,\phi)$, where
\begin{itemize}
\item[(i)] $M=\{M_q\}$ generates a chain complex, denoted by ${\cal M}$,
with differential $d'$ such that any non-null entry of SNF of
  the matrix of $d'$ in each dim $q$,
 is greater than $1$.
\item[(ii)]  $(f,g,\phi)$ defines a chain contraction of ${\cal C}$ to ${\cal M}$.
  \end{itemize}

\begin{figure}[t!]
\begin{center}\includegraphics[width=8cm]{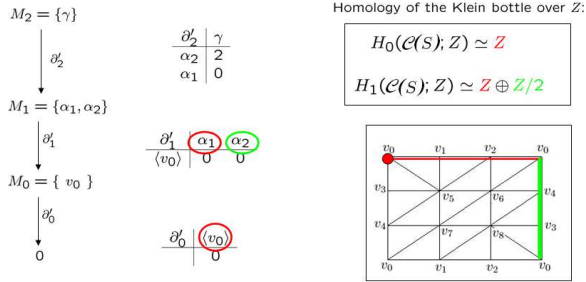}\end{center}
\caption{Homological information obtained from an AM-model for the Klein bottle.}
\label{4}\end{figure}

Property (ii)  follows that  $H({\cal C};{\bf Z})$ and
$H({\cal M};{\bf Z})$ have isomorphic homology groups. Moreover,
integer (co)homology generators and representative (co)\-cycles of (co)homology generators
 of ${\cal C}$
 can directly be obtained from ${\cal M}$. See Fig. \ref{4}, for  example, where $M_1=\{\alpha,\beta\}$, $\alpha$ being a representative cycle of a generator of the free part of $H_1({\cal C}(S);{\bf Z})$ and $\beta$ 
 a representative cycle of a generator of the torsion part.
An algorithm for computing AM-models is given in \cite{GR06,GR08}. It
needs to reduce the matrices of the differential to
its SNF. 

An {\em algebraic topological model} (AT-model) \cite{GR03,GR05}
is a  tool for computing over a field ${\bf Z}/p$, $p$ being a
prime, (co)\-homology, representative (co)cycles of (co)\-homology
generators and the cup product  on cohomology of nD digital
images (and hence also the cohomological invariant $HB1$). The algorithm presented in \cite{GR03,GR05} for
computing AT-models over fields
 runs in time at most ${\cal O}(m^3)$, where $m$ is the number
of generators of the given chain complex.
More concretely, an \textit{AT-model}
 for a chain complex ${\cal C}=(C,d)$ is a set
 $((C,d), H, f,g,\phi)$,
 where
 \begin{itemize}
 \item[(i)] $H=\{H_q\}$ generates  a chain complex, denoted by
 ${\cal H}$, with null differential.
\item[(ii)] $(f,g,\phi)$ defines a chain contraction of ${\cal C}$ to ${\cal H}$ over ${\bf Z}/p$.
\end{itemize}
Property (ii) follows that the homology groups of ${\cal C}$
  and ${\cal H}$, with coefficients
over ${\bf Z}/p$, are isomorphic.

\begin{example}
Let $K$ be the simplicial complex shown in Fig. \ref{5}. The set $((K,\partial),$ $H_{\scst K},f_{\scst K},g_{\scst K},\phi_{\scst K})$ is
an AT-model for $K$ over ${\bf Z}/2$, where
$$K=\{v_1, v_2, v_3, v_4, v_1v_4, v_1v_2, v_2v_4, v_3v_4, v_2v_3, v_1v_2v_4\},\;
H_{\scst K}=\{v_4, v_2v_3\}$$
and the images of the maps $f_{\scst K}$, $g_{\scst k}$ and $\phi_{\scst K}$ are:

\begin{center}
\begin{tabular}{c|c|c|c|c|c|c|c|c|c|c}
& $v_1$ & $v_2$ & $v_3$ & $v_4$ & $v_1v_4$ & $v_1v_2$ & $v_2v_4$ & $v_3v_4$ & $v_2v_3$ & $v_1v_2v_4$\\
\hline
$f_{\scst K}$ & $v_4$ & $v_4$ & $v_4$ & $v_4$ & $0$ & $0$ & $0$ & $0$ & $v_2v_3$ & $0$\\
$g_{\scst K}$  &  &  & $v_4$ &  &  &  &  & $\alpha$ & \\
$\phi_{\scst K}$ & $v_1v_4$ & $v_1v_2+v_1v_4$ & $v_3v_4$ & $0$ & $0$ & $v_1v_2v_4$ & $0$ & $0$ & $0$ & $0$\\
\end{tabular}
\end{center}

where $\alpha=v_2v_3+v_2v_4+v_3v_4$. Then, the homology groups of ${\cal C}(K)$ over the field ${\bf Z}/2$ are $H_0({\cal C}(K);{\bf Z}/2)={\bf Z}/2$, $H_1({\cal C}(K);{\bf Z}/2)={\bf Z}/2$ and $H_2({\cal C}(K);{\bf Z}/2)=0$. The representative cycles of homology generators in dim $0$ and $1$ are $v_4$ and $\alpha$, respectively.

\begin{figure}[t!]
\begin{center}\includegraphics[width=8cm]{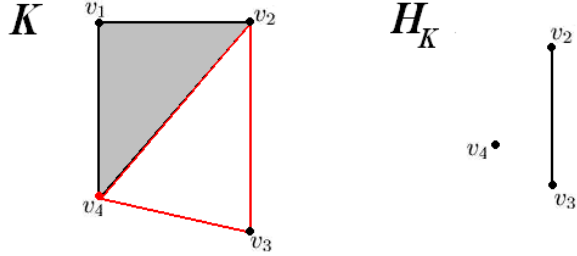}\end{center}
\caption{On the left, the simplicial complex $K$; in red, the representative
cycles of homology generators of $H({\cal C}(K);{\bf Z}/2)$. On the right, the elements of $H_{\scst K}$.}
\label{5}\end{figure}

\end{example}

The main problem of this tool
is that from the homology obtained from an AT-model,  it is not possible, in general, to get all 
the torsion information of the integer homology of the object under study (see Example \ref{toroklein}).

\section{The Notion of $\lambda$-AT-Model}
As far as we know all the algorithms in the literature for computing integer homology (with or without torsion)
need to reduce the matrix of the differential to
its SNF.
Our aim is  to compute integer homological information
avoiding the computation of SNF. For this task,
 we first define the notion of
$\lambda$-AT-model  which can be seen  as a
 generalization of the concept of AT-model, working in the integer domain.
We study its properties and give an algorithm for computing it. Moreover,
we will see that from a $\lambda$-AT-model for a chain complex ${\cal C}$, we can
 obtain the Betti numbers $\beta_q$, the prime numbers  $p$ involved in the invariant factors of
 the torsion subgroup of $H({\cal C};{\bf Z})$,
 the amount of invariant factors that are a power of $p$, and a set of representative cycles
 of generators of homology mod $p$, for each $p$. That is, from a $\lambda$-AT-model for a chain complex ${\cal C}$, we can obtain a set ${\cal I}$ of independent non-boundary cycles such that given any cycle $c$ of ${\cal C}$, $\lambda c$ can be expressed as a linear combination over ${\bf Z}$ of the cycles of ${\cal I}$. Finally, we  present an algorithm for extracting this homological information, performed in ${\cal O}(m^3 \psi(\lambda))$ in the worst case, $\psi$ being the Euler function.

First of all, let us define the concept of $\lambda$-chain contraction as a generalization of chain
contraction and
study its properties. Later, we will define a $\lambda$-AT-model as a $\lambda$-chain contraction of a chain complex ${\cal C}$
to another chain complex with null differential.

\begin{definition}
Let $\lambda$ be an integer, $\lambda\neq 0$.
A $\lambda$-\textit{chain contraction} of a chain complex ${\cal C}$,
 to
another chain complex ${\cal C'}$ with differentials $d$ and $d'$, respectively, is a set
$(f,g,\phi,\lambda)$ such that: $f : {\cal C} \rightarrow {\cal C'}$ and
$g: {\cal C'} \rightarrow {\cal C}$ are chain maps; $fg$ is $\lambda id_{\scst \cal C'}$
and $\phi: {\cal C} \rightarrow {\cal
C}$ is a chain homotopy of $\lambda id_{\scst \cal C'}$ to
$gf$, that is,
$$fg=\lambda id_{\scst \cal C'}\qquad\mbox{ and }\qquad
 \lambda id_{\scst \cal C} - gf=\phi d + d \phi\,.$$
\end{definition}

Observe that a $\lambda$-chain contraction with $\lambda=1$ is a chain contraction.
Moreover, if $(f,g,\phi,\lambda)$ is a $\lambda$-chain contraction of ${\cal C}$ to ${\cal C'}$ with $\lambda=-1$ then
$(-f,g,-\phi)$ is a chain contraction of ${\cal C}$ to ${\cal C'}$.

\begin{lemma}\label{truco}
If $((C,d),H,f,g,\phi,\lambda)$ is a $\lambda$-AT-model,  with  $\lambda<0$, 
then $((C,d),$ $H,-f,g,-\phi,-\lambda)$ is a $\lambda'$-AT-model,  with  $\lambda':=-\lambda>0$. 
\end{lemma}

If $(f,g,\phi,\lambda)$ is a $\lambda$-chain contraction of ${\cal C}$ to ${\cal C'}$ and
$(f',g',\phi',\lambda')$ is a $\lambda'$-chain contraction of ${\cal C'}$ to ${\cal C''}$, then
\begin{eqnarray}
\label{composition}(f'f,gg', \lambda'\phi+g\phi'f,\lambda\lambda')\end{eqnarray}
is a $\lambda\lambda'$-chain contraction of ${\cal C}$ to ${\cal C''}$.
Another important property is the following. \label{lchain}

\begin{prop}\label{free}
If there exists a $\lambda$-chain contraction $(f,g,\phi,\lambda)$ (with $\lambda\neq 0$) of a chain complex
${\cal C}$ to another
chain complex ${\cal C'}$ then the free subgroups of
$H({\cal C};{\bf Z})$ and $H({\cal C'};{\bf Z})$ are isomorphic. As a consequence,
the Betti numbers of ${\cal C}$ and ${\cal C'}$ coincide.
\end{prop}

\begin{pf}
 Let $d$ and $d'$ be the differentials of ${\cal C}$ and ${\cal
C'}$, respectively. Let $a$ be a representative cycle of a
homology generator of the free subgroup of
 $H({\cal C};{\bf Z})$, that is, $d(a)=0$ and for each
$\mu \in {\bf Z}$, $\mu \neq 0$, $\mu a \notin Im \, d$. Then
$f(a)$ is a cycle of ${\cal C'}$ since $f$ is a chain map
and therefore $d'f(a)=fd(a)=0$. Suppose that  there exists an integer $\mu\neq
0$ and a chain $b'\in {\cal C'}$ such that $\mu f(a)=d'(b')$.
Then,  $\lambda a -  gf(a)=d
\phi(a)$ since $(f,g,\phi,\lambda)$ is a $\lambda$-chain
contraction and $a$ is a cycle. Therefore, $\lambda\mu
a=d(g(b')+\mu\phi(a))$ which is a contradiction since $a$ is a
representative cycle of a homology generator of the free subgroup
of $H({\cal C};{\bf Z})$. Therefore $f(a)$ is a representative
cycle of a homology generator of the free subgroup of $H({\cal
C'};{\bf Z})$.

Now, let $a$ and $b$ be representative cycles of two different
homology generators of the free subgroup of $H({\cal C};{\bf
Z})$. Suppose that $f(a)$ and $f(b)$ are representative cycles of
the same homology generator of the free subgroup of $H({\cal
C'};{\bf Z})$. Then there exists a chain $b'\in {\cal C'}$ such
that $f(a)-f(b)=d'(b')$. Then $\lambda(a-b)=d(g(b')+\phi(a-b))$
which is a contradiction. Therefore,  $f(a)$ and $f(b)$ are representative
cycles of two different homology generators of the free subgroup
of $H({\cal C'};{\bf Z})$. Analogously, it is true that if
$a'$ and $b'$ are representative cycles of two different homology
generators of the free subgroup of
 $H({\cal C'};{\bf Z})$ then $g(a')$ and $g(b')$ are
representative cycles of two different homology generators of the
free subgroup of $H({\cal C};{\bf Z})$. We can conclude that the
free subgroup of $H({\cal C};{\bf Z})$ and $H({\cal C'};{\bf Z})$
are isomorphic.
\qed
\end{pf}

Now, define a $\lambda$-AT-model for a chain complex ${\cal C}$
as a  $\lambda$-chain contraction of ${\cal C}$ to a chain complex with null differential.

\begin{definition}
 Let ${\cal C}=(C,d))$ be a chain complex.
 Let $\lambda$ be  a non-null integer.
A $\lambda$-\textit{AT-model} for ${\cal C}$
is a set
 $((C,d), H, f,g,\phi,\lambda)$,
 where
 \begin{itemize}
 \item[(i)] $H=\{H_q\}$ generates  a chain complex, denoted by ${\cal H}$, with null differential.
\item[(ii)] $(f,g,\phi,\lambda)$ defines a $\lambda$-chain contraction of ${\cal C}$ to ${\cal H}$.
 \end{itemize}
\end{definition}

As a consequence of Prop. \ref{free} and since the differential of ${\cal H}$ is null,
if there exists a $\lambda$-AT-model $((C,d), H, f,g,\phi,\lambda)$ for ${\cal C}$
then the free subgroup of $H({\cal C};{\bf Z})$ is isomorphic to ${\cal H}$.

\begin{prop} Given a $\lambda$-AT-model,  a rational AT-model (i.e., an AT-model over ${\bf Q}$)
can  directly be obtained
as well as
rational (co)homology and representative (co)cycles of (co)homology generators. Concretely, if
 $((C,d),H,$ $f,g,\phi,\lambda)$ is a $\lambda$-AT-model for a chain complex ${\cal C}$, then
$((C,d),H,\frac{1}{\lambda} f,g,\frac{1}{\lambda} \phi)$
is an AT-model for ${\cal C}$
over ${\bf Q}$ and $\{g(h): h\in H\}$ is a set of representative cycles of generators of
$H({\cal C};{\bf Q})$.
\end{prop}

Universal Coefficient Theorem for Homology \cite[pp. 332]{Munkres} states that each Betti number $\beta_q$
coincides with the vector space dimension of $H_q({\cal C};{\bf Q})$.
In addition, working  in the integer domain, a $\lambda$-AT-model for ${\cal C}$ provides
a set of $\beta_q$ independent non-boundary $q$-cycles of ${\cal C}$ over ${\bf Z}$, for each $q$.

\begin{cor}\label{coro1}
Let $((C,d),H,f,g,\phi,\lambda)$ be a $\lambda$-AT-model for a given chain complex ${\cal C}$.  Then
${\cal H}$ is isomorphic to the free subgroup of $H({\cal C};{\bf Z})$.
Moreover, the set $\{g(h):\, h\in H\}$
is a set of independent non-boundary cycles of ${\cal C}$ over ${\bf Z}$ that generate the set $\{\lambda\alpha:\, \alpha\in F({\cal C};{\bf Z})\}$.
\end{cor}

\begin{pf}
Let $((C,d),H,f,g,\phi,\lambda)$ be a $\lambda$-AT-model for ${\cal C}$ and $\{h_1, \dots, h_m\}$ a set of generators of $H$.
Suppose that $\{g(h_1),\dots,g(h_m)\}$ are not independent. Then there exists $i$, $1\leq i\leq m$, such that
$g(h_i)=\sum_{j=1}^m c_j g(h_j)$ for some $c_j\in{\bf Z}$, $1\leq j\neq i\leq m$, and $c_i=0$.
Therefore 
$f g(h_i)=\sum_{j=1}^m c_j fg(h_j)$, so that 
$\lambda h_i=\sum_{j=1}^m c_j \lambda h_j$. Simplifying, $ h_i=\sum_{j=1}^m c_j h_j$, which is a contradiction.
 Let $\alpha$ be a homology class of the free part of $H({\cal C};{\bf Z})$ and $c$ a representative cycle of $\alpha$, that is, $\alpha=[c]$. Since $f(c)\in H$, then $f(c)=\sum_{i=1}^m c_i h_i$, $c_i \in {\bf Z}$. Therefore, $g f(c)=\sum_{i=1}^m c_i g(h_i)$. So, 
$\sum_{i=1}^m c_i g(h_i)=\lambda c - \phi d (c) - d \phi(c)$. Since $c$ is a cycle, then $\sum_{i=1}^m c_i g(h_i)=\lambda c - d \phi(c)$. Considering homology classes, $\sum_{i=1}^m c_i [g(h_i)]=\lambda [c]$. Consequently, $\{g(h_1), \dots, g(h_n)\}$ is a set of independent non-boundary representative cycles of generators that generate the set $\{\lambda \alpha \, : \, \alpha \in F({\cal C};{\bf Z})\}$.
\qed
\end{pf}

Similar to rational AT-models, an AT-model over ${\bf Z}/p$ (where
$p$ is a prime which does not divide $\lambda$) can always be defined from a $\lambda$-AT-model.

\begin{prop}
Given a $\lambda$-AT-model $((C,d),H,f,g,\phi,\lambda)$ and a prime $p$ such that $p$ does not divide $\lambda$, then
$((C,d_{\scst {\bf Z}/p}), H,f_{\scst {\bf Z}/p},g_{\scst {\bf Z}/p},\phi_{\scst {\bf Z}/p})$,
where
$$d_{\scst {\bf Z}/p}=d\mbox{ mod }p,\;\;
f_{\scst {\bf Z}/p}=\lambda^{-1}f\mbox{ mod }p,\;\;
g_{\scst {\bf Z}/p}=g\mbox{ mod }p,\;\;
\phi_{\scst {\bf Z}/p}=\lambda^{-1}\phi\mbox{ mod }p,$$
is an AT-model for ${\cal C}$ over ${\bf Z}/p$. Moreover,
$\{g_{\scst {\bf Z}/p}(h):h\in H\}$ is a set of representative cycles of generators of $H({\cal C};{\bf Z}/p)$.
\end{prop}

An important property of $\lambda$-AT-models is that we can obtain the prime numbers $p$ involved in the torsion subgroup
of the homology of ${\cal C}$.

\begin{prop}\label{fcero}
Let $((C,d),H,f,g,\phi,\lambda)$ be a $\lambda$-AT-model.  Let
 $a\in {\cal C}$ such that $d(a)=0$.
If there exists
$b\in {\cal C}$ such that
$d(b)=\mu  a$ where  $\mu\in {\bf Z}$,
$\mu\neq 0$  and
for each $\rho$, where $0<\rho<\mu$,  $\rho a\not\in$ Im $d$, then
$f(a)=0$ and
  $\mu$ divides $\lambda$.
\end{prop}

\begin{pf}
   Suppose that
$b\in {\cal C}$ such that
$d(b)=\mu  a$ where  $\mu\in {\bf Z}$,
$\mu\neq 0$ and for each $\rho$, where $0<\rho<\mu$,  $\rho a\not\in$ Im $d$.
First, let us prove that $f(a)=0$. 
Suppose that $f(a)\neq 0$. Then  $\mu f(a)\neq 0$ (since the ground ring is  ${\bf Z}$).
On the other hand,  $\mu f(a)=f(\mu  a)=f(d(b))=0$, since $f$ is a chain map and the differential of
${\cal H}$ is null. We then have that $f(a)=0$ which is a contradiction.
Now, suppose that $\mu$ does not divide $\lambda$, then there exist $s,r\in {\bf Z}$,
such that
$0<r<\mu$ and $\lambda=s \mu+r$. On one hand,  $r a\not\in$ Im $d$.
On  the other hand,
 $r a=(\lambda -s \mu)a=
 \lambda a-s\mu a=gf(a)+\phi d(a)+d\phi(a)-s d(b)=d(\phi(a)-sb)\in$ Im $d$ which is a contradiction.
We conclude that $\mu$ divides $\lambda$.
\qed
\end{pf}

\begin{cor}\label{coro2} Let $((C,d),H,f,g,\phi,\lambda)$ be a $\lambda$-AT-model.
If  $p^{t_{(q,p)}}$ is an invariant factor of $H_q({\cal C};{\bf Z})$,
then $p$ divides $\lambda$.
\end{cor}

Observe that if $\lambda$ is `large' then there are lots of candidates to take part in the torsion subgroup
of $H({\cal C};{\bf Z})$ (all the primes that divide $\lambda$). Therefore, an important
task could be to get the minimum $\lambda>0$ such that there exists a $\lambda$-AT-model for ${\cal C}$.

\begin{prop}
Let ${\cal C}=(C,d)$ be a chain complex such that in each dim $q$ the matrix of $d_q$
 coincides with its SNF. Let $\rho$ be the lowest common multiple of
all the non null elements that appear in the matrices of $d_q$ for all $q$. Define
the homomorphisms $f$, $g$ and $\phi$ and the set $H$ as follows:
\begin{itemize}
\item If $x\in C$  such that $d(x)=\mu y$ for some integer $\mu>0$ then
$$f(x)=0\,,\;f(y)=0\,,\; \phi(x)=0\;\mbox{  and }\;\phi(y)=\frac{\rho}{\mu}\,x\,.$$
\item If $x\in C$ such that $d(x)=0$ and there is not any $z\in C$ being
$d(z)=\mu x$ for some integer $\mu>0$, then
$$f(x)=x\,, \;g(x)=\rho x\;\mbox{  and }\;\phi(x)=0\,.$$
\item  $H=\{x\in C$ such that $d(x)=0$ and there is not any $z\in C$ such that
$d(z)=\mu x$ for some integer $\mu>0\}$.
\end{itemize}
The set $((C,d),H,f,g,\phi,\rho)$ defines a $\rho$-AT-model. Moreover, this
$\rho$-AT-model satisfies that a prime $p$ divides $\rho$ if and only if it takes part in the torsion subgroup
of $H({\cal C};{\bf Z})$.
\end{prop}

Observe that in order to obtain a $\lambda$-AT-model for a chain complex ${\cal C}$ with minimum $\lambda$,
we need to compute the SNF of the matrix of the differential of ${\cal C}$ in each dimension.
The following algorithm computes a $\lambda$-AT-model for a chain complex ${\cal C}$
without computing the SNF of the differential. In this case, $\lambda$ might be non-minimum.

\begin{alg}\label{latmodel}
Computing a $\lambda$-AT-model for a chain complex ${\cal C}$.

\begin{tabbing}
{\sc Input:} \= {\tt A chain complex ${\cal C}=(C,d)$ of dim $n$.}\\\\
 {\tt $H_0:=C_0$, $A:=C_0$, $\lambda:=1$, $f_0:=id_{C_0}$, $g_0:=id_{H_0}$,
   $\phi_0:=0$.} \\
   {\tt For}\= { \tt $q=1$ to $q=n$ do}\\
\> {\tt $H_q:=\{ \, \}$, $x:=0$, $\gamma:=0$, $\gamma':=0$.}\\
\> {\tt While}\= { \tt $M:=\{|\langle f_{q-1}d_q(c), c' \rangle|\neq 0:\, c\in C_q,\, c'\in H_{q-1}
\}$ is not empty do}\\
\>\> {\tt Take}\={ \tt $\alpha\in H_{q-1}, \;\beta\in C_{q}$ s.t.
$|\langle f_{q-1}d_q(\beta), \alpha \rangle|=$min $M$ then}\\
\>\>\>{\tt $H_{q-1}:=H_{q-1} \setminus \{\alpha\},\; A:=A \cup \{\beta\}$, $f_q(\beta):=0$, $\phi_q(\beta):=0$,}\\
\>\>\> {\tt $x:=\langle f_{q-1}d_q(\beta),\alpha \rangle,\;
\gamma:=f_{q-1}d_q(\beta), \;
\gamma':=\lambda \beta - \phi_{q-1}d_q(\beta)$.}\\
\>\>\> {\tt For} \= { \tt each $b \in C_r$, $0 \leq r \leq q-1$, do}\\
\>\>\>\>{\tt $f_r(b):=x f_r(b) -
\langle f_r (b), \alpha \rangle \gamma$,}\\
\>\>\>\>{\tt $\phi_r(b):=x \phi_r(b) +
\langle f_r (b), \alpha \rangle \gamma'$,}\\
\>\>\>{\tt $\lambda:=x \lambda$.}\\
\>{\tt For}\= { \tt each $a \in C_q\setminus A$ do}\\
\>\> {\tt $H_q:=H_q \cup \{a\},\; A:=A\cup\{a\}$,}\\
\>\>{\tt  $f_q(a):=a,\; g_q(a):=\lambda a - \phi_{q-1}d_q(a),\; \phi_q(a):=0$.}\\
\\
\textsc{Output:} \= {\tt The set
$((C,d),H,f,g,\phi,\lambda)$.}
\end{tabbing}
\end{alg}

Fixed $q$, there are two different parts in the algorithm. First,  
while the matrix $M$ corresponding to $f_{q-1}d_q$ is non-null, take $\alpha\in H_{q-1}$ and $\beta\in C_q$ such that 
$|\langle f_{q-1}d_q(\beta),\alpha\rangle|$ is the minimum of $M$ (without considering signs). Then, $\alpha$ is deleted from $H_{q-1}$ (i.e., a homology class is destroyed) and $\beta$ is added to $A$ (i.e, $\beta$ has been used). 
Second, if the new redefined matrix $M$ is null,  add, to $H_q$, the non-used elements of $C_q$ (i.e. new classes of homology are created).
At the end of the algorithm, the set $A$ consists in an ordered set of all the elements of $C$. 

\begin{thm}\label{thlatmodel}
The set $((C,d ),H,f,g,\phi,\lambda)$  obtained applying Alg. \ref{latmodel}
defines a $\lambda$-AT-model for the chain complex ${\cal C}=(C,d)$.
\end{thm}

\begin{pf} 
Let $A=\{a^1,\dots,a^m\}$ such that if $i<j$ then $a^i$ was added to $A$ before $a^j$ when the algorithm was applied.  
Fixed $i$,   
assume that $f_{r-1}d_r(a^k)=0$, $\lambda a^k - g_r f_r (a^k)=\phi_{r-1}d_r(a^k)+
d_{r+1}\phi_r(a^k)$, 
 $f_{r+1} \phi_r(a^k)=0$ and,  if $a^k\in H_r$,
 $d_r g_r(a^k)=0$ and
$f_r g_r(a^k)=\lambda a^k$,  for $1\leq k<i$ and $r=dim (a^k)$.
 
Let $q=dim(a^i)$. We have to prove that the redefined maps that we will denote here by 
 $f'$, $g'$ and $\phi'$ and the new integer $\lambda '$ satisfy that 
 $f'_{r-1}d_r(a^k)=0$,
 $\lambda' a^k-g'_r f'_r(a^k)=\phi'_{r-1}d_r(a^k)+d_{r+1} \phi'_r(a^k)$,
  $f'_{r+1} \phi'_r(a^k)=0$ and, if $a^k\in H_r$, 
 $d_r g'_r(a^k)=0$ and 
 $f'_r g'_r(a^k)=\lambda' a^k$    for $1\leq k\leq i$ and $r=dim(a^k)$. 

We have to consider two cases. (1) If $a^i$ was added to $A$ while $M$ was non-empty, then  $g'=g$ (that is, $g$ does not change). 
Therefore, if $a^k\in H_r$ for some $r$, then $d_r g'_r(a^k)=d_r g_r(a^k)=0$ for all $k$, $1\leq k<i$.
Consider the element $\alpha\in H_{q-1}$ that defines
the auxiliary parameter $x=\langle f_{q-1}d_q(a^i),\alpha\rangle$ at this stage. 

If $k=i$, then   $a^i\not\in H_q$. We have that 
$f'_{q-1}d_q(a^i)=x f_{q-1}d_q(a^i)-\langle f_{q-1}d_q(a^i),\alpha\rangle \gamma$ $=x\gamma-x\gamma=0$ and
  $\phi'_{q-1}d_q(a^i)+d_{q+1} \phi'_q(a^i)=\phi'_{q-1}d_q(a^i)=x \phi_{q-1}d_q(a^i)+\langle f_{q-1}d_q(a^i),$ $\alpha\rangle \gamma'=
 x (\phi_{q-1}d_q(a^i)+\lambda a^i-\phi_{q-1}d_q(a^i))=x\lambda a^i=\lambda' a^i$.
  Moreover, 
 $f'_{q+1} \phi'_q(a^i)=0$ since $\phi'_q(a^i)=0$.

Observe that:  For any $r$-chain $b$ such that $r< q-1$, $f'_r(b)=x f_r(b)$ and $\phi'_r(b)=x \phi_r(b)$.   $d_q(\gamma')=d_q(\lambda a^i-\phi_{q-1}d_q(a^i))=\lambda d_q(a^i) -(\lambda d_q(a^i)-g_{q-1}f_{q-1}d_q(a^i)-\phi_{q-1} d_{q-1}d_q(a^i))=g_{q-1}f_{q-1}d_q(a^i)=
g_{q-1}(\gamma)$.  For any $q$-chain $b$, $f_q(b)=0$ 
 and $\phi_q(b)=0$, therefore, $f'_q(b)=0$ and $\phi'_q(b)=0$,
 since $a^i$ was added to $A$ while $M$ was non-empty.

For $1\leq k<i$. (1.a) If $r:=dim(a^k)<q-1$ then 
$f'_{r-1}d_r(a^k)=x f_{r-1}d_r(a^k)=0$ by induction and
$\phi'_{r-1}d_r(a^k)+d_{r+1}\phi'_r(a^k)=x(\phi_{r-1}d_r(a^k)+d_{r+1}\phi_r(a^k))=
x(\lambda a^k-g_r f_r(a^k))=\lambda' a^k-g_r f'_r(a^k)$.
Moreover, $f'_{r+1}\phi'_r(a^k)=x f'_{r+1}\phi_r(a^k)=x(x f_{r+1}\phi_r(a^k)-\langle f_{r+1}\phi_r(a^k),\alpha\rangle\gamma)
=0$ by induction.
 If $a^k\in H_r$ then 
  $f'_r g_r(a^k)=x f_r g_r(a^k)=x\lambda a^k=\lambda' a^k$.
(1.b) If $dim (a^k)=q-1$, then 
$f'_{q-2}d_{q-1}(a^k)=x f_{q-2}d_{q-1}(a^k)=0$ by induction and
$\phi'_{q-2}d_{q-1}(a^k)+d_{q}\phi'_{q-1}(a^k)=x \phi_{q-2}d_{q-1}(a^k)$ $+d_q(x\phi_{q-1}(a^k)+\langle f_{q-1}(a^k), \alpha\rangle\gamma')=x(\phi_{q-2}d_{q-1}(a^k)+d_q\phi_{q-1}(a^k))+\langle f_{q-1}(a^k),$ $\alpha\rangle d_q(\gamma')
=x(\lambda a^k-g_{q-1}f_{q-1}(a^k))+\langle f_{q-1}(a^k),\alpha\rangle g_{q-1}(\gamma)
=\lambda'a^k-g_{q-1}f'_{q-1}(a^k)$. 
Moreover, $f'_{q}\phi'_{q-1}(a^k)=
f'_q(x\phi_{q-1}($ $a^k)+\langle f_{q-1}(a^k),\alpha\rangle \gamma')$. 
Now, $f'_q \phi_{q-1}(a^k)=0$ since for any $q$-chain $b$, $f'_q(b)=0$;
$f'_q(\gamma')=f'_q(\lambda a^i-\phi_{q-1}d_q(a^i) )=0$ for the same reason than above. Then, $f'_{q}\phi'_{q-1}(a^k)=0$. 
If $a^k\in H_{q-1}$ then $a^k\neq \alpha$ and  $f'_{q-1} g_{q-1}(a^k)=x f_{q-1} g_{q-1}(a^k)-
\langle f_{q-1}g_{q-1}(a^k),\alpha\rangle \gamma=x \lambda a^k-\langle \lambda a^k,\alpha\rangle \gamma=
x \lambda a^k=\lambda' a^k$.
(1.c) If $dim (a^k)=q$, then 
$f'_{q-1}d_q(a^k)=x f_{q-1}d_q(a^k)-\langle f_{q-1}d_q(a^k),$ $\alpha\rangle\gamma=0$ by induction and
$\phi'_{q-1}d_q(a^k)+d_{q+1}\phi'_q(a^k)=x(\phi_{q-1}d_q(a^k)+d_{q+1}\phi_q(a^k))$ $
+\langle f_{q-1}d_q(a^k),\alpha\rangle\gamma'
=
x(\lambda a^k-g_q f_q(a^k))=\lambda' a^k-g_q f'_q(a^k)$.
Moreover, $f'_{q+1}\phi'_q(a^k)$ $=0$.

(2) If $a^i$ was added to $A$ when $M$ was empty, then $\lambda'=\lambda$ (that is, $\lambda$ does not change).
First,   $f'_{q-1} d_q(a^i)=f_{q-1}d_q(a^i)=0$ by induction and $\lambda a^i-g'_q f'_q(a^i)=\lambda a^i- g'_q(a^i)=
\lambda a^i-(\lambda a^i-\phi_{q-1}d_q(a^i))=\phi_{q-1}d_q(a^i)=\phi'_{q-1}d_q(a^i)=\phi'_{q-1}d_q(a^i)
+d_{q+1}\phi'_q(a^i)$. Second, $d_q g'_q(a^i)=\lambda d_q(a^i)-d_q\phi_{q-1}d_q(a^i)=\lambda d_q(a^i)-(\lambda d_q(a^i)-g_{q-1} f_{q-1} d_q(a^i)-\phi_{q_2} d_{q-1}  d_q(a^i))=g_{q-1} f_{q-1}d_q(a^i)=0$  and $f'_q g'_q(a^i)=f'_q(\lambda a^i-\phi_{q-1}d_q(a^i))=
\lambda a^i-f_q\phi_{q-1}d_q(a^i)=\lambda a^i$,
since  $f_{r+1} \phi_r(a^k)=0$  for $1\leq k< i$ and $r=dim(a^k)$, by induction. 
Finally, $f'_{q+1}\phi'_q(a^i)=0$ since $\phi'_q(a^i)=0$.
\qed
\end{pf}

To study the complexity, consider the order of the elements  of $C$ given by the set $A$. Fixed $i$,  
count the number of elementary operations involved.
We have  to update $f_{r}(b)$ and $\phi_{r}(b)$ for each $b\in C_{r}$, $0\leq r\leq q-1$.
Observe that in order to update $f_{r}(b)$ and $\phi_{r}(b)$ in the worst case, we have to add to both a chain containing $m_q$ generators (if $m_q$ is the number of generators of dim $q$). In the worst case, $m_q$ could be of the same order than $m$ (the total number of generators of ${\cal C}$).
So, fixed $i$, the total cost of these operations is $O(m^2)$.
Therefore, the total algorithm runs in time at most $O(m^3)$.

\begin{figure}[t!]
\begin{center}\includegraphics[width=8cm]{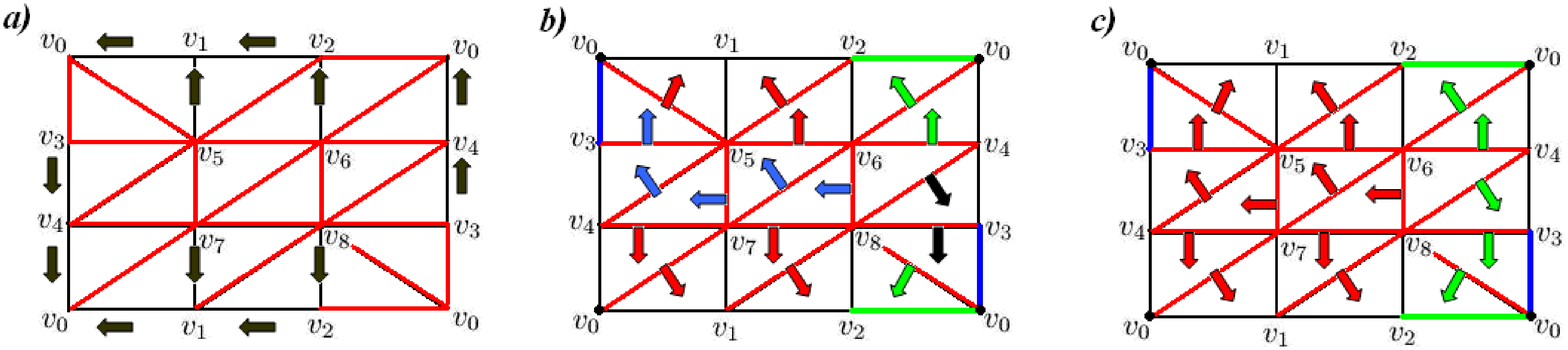}\end{center}
\label{6}\caption{ A visual interpretation of the maps $f_{\scst S}$ and $\phi_{\scst S}$ on a triangulation of the Klein bottle when applying Alg. \ref{latmodel}: a) after adding all the vertices and edges; b) after adding all the vertices edges and triangles except for $v_4v_6v_8$; c) at the end.}
\end{figure}

\begin{example}\label{exklein}
Consider the simplicial complex $S$ derived from the triangulation
of the Klein bottle given in Fig. \ref{6} and the chain complex
${\cal C}(S)$ of dim $2$ associated to $S$. Applying Alg. \ref{latmodel},
we obtain a $2$-AT-model, $(C(S),H_{\scst S},f_{\scst S},g_{\scst S},$ $\phi_{\scst S},2)$, for ${\cal C}(S)$. 
For the sake of simplicity in this example, each time a negative value for $\lambda$ appears, we apply Lemma \ref{truco} and turn it into positive. 
After running Alg. \ref{latmodel} only for the vertices and edges of $S$, 
we get that, at this stage,  $H_{{\scst S}}=\{v_0\}\cup \{x:\; x$ is an edge marked in red in Fig. \ref{6}.a)$\}$ and the values of $f_{\scst S}$, $g_{\scst S}$ and $\phi_{\scst S}$ at this stage are:

\begin{center}
\begin{tabular}{c|c|c|c|c}
& $v_0$ & $v_i$ & $y$ &$x$  \\
\hline
$f_{\scst S}$ & $v_0$ & $v_0$ & $0$ & $x$ \\
$g_{\scst S}$ & $v_0$  & &  & $x-\phi_{{\scst S}}\partial_1(x)$\\
$\phi_{\scst S}$ & $0$ & $\gamma_{(v_0,v_i)}$ & $0$ & $0$\\
\end{tabular}
\end{center}

\noindent where $1 \leq i \leq 8$,  $y$ denotes an edge drawn in black in Fig. \ref{6}.a) and $\gamma_{(v_0,v_i)}$ is a path connecting the vertices $v_0$ and $v_i$, $i \neq 0$ following the arrows drawn in black in Fig. \ref{6}.a); for example, $\gamma_{(v_0,v_6)}=v_0v_1
+  v_1v_2 + v_2v_6$. 

Now, apply the algorithm for all the vertices, edges and triangles except for the triangle $v_4v_6v_8$ (to get a better idea of the geometric meaning of the maps $f_{\scst S}$, $g_{\scst S}$ and $\phi_{\scst S}$, suppose that we do not consider  this triangle until the end). We get that $H_{{\scst S}}=\{v_0,v_0v_2,v_0v_3\}$. 
The values of $f_{\scst S}$, $g_{\scst S}$ and $\phi_{\scst S}$ at this stage are:

\begin{center}
\begin{tabular}{c|c|c|c|c|c|c|c}
& $v_0$ & $v_i$ & $y$ & $v_0v_2$ & $v_0v_3$ & $z$ & $t$\\
\hline
$f_{\scst S}$ & $v_0$ & $v_0$ & $0$ & $ v_0v_2$ & $v_0v_3$ & $\gamma'_{(z)}$ & $0$\\
$g_{\scst S}$ & $v_0$ & & & $v_0v_2-v_1v_2-v_0v_1$& $v_0v_3+v_3v_4-v_0v_4$ &  &\\
$\phi_{\scst S}$ & $0$  & $\gamma_{(v_0,v_i)}$ & $0$ & $0$ & $0$ & $\gamma''_{(z)}$ & $0$\\
\end{tabular}
\end{center}

\noindent where $z$ denotes one red edge (see  Fig. \ref{6}.b)) and $t$ a triangle except for $v_4v_6v_8$.
$\gamma'_{z}$ is $0$, $\pm v_0v_3$, $\pm v_0v_2$ or $v_0v_2-v_0v_3$
if the arrow drawn over $z$ in Fig. \ref{6}.b) is, respectively, red, blue, green or black.  
$\gamma''_{(z)}$
is a `path' from $z$ following the arrows drawn in Fig. \ref{6}.b).
For example,
$\gamma_{(v_6v_7 )}=v_5v_6v_7 -v_2v_5v_6+v_1v_2v_5+v_4v_5v_7-v_3v_4v_5-v_0v_3v_5+v_0v_1 v_5+v_0v_4v_7-v_0v_1v_7$.

Now, add the triangle $v_4v_6v_8$. Then $f_{\scst S}\partial(v_4v_6v_8)$ is $2v_0v_3$. Therefore, the new images for $f_{\scst S}$,
$g_{\scst S }$ and $\phi_{\scst S}$ are:

\begin{center}
\begin{tabular}{c|c|c|c|c|c|c|c|c}
& $v_0$ & $v_i$ & $y$ & $v_0v_2$ & $v_0v_3$ & $z$ & $t$ & $v_4v_6v_8$\\
\hline
$f_{\scst S}$ & $2v_0$ & $2v_0$ & $0$ & $ 2v_0v_2$ & 0 & $2\gamma'''_{(z)}$ & $0$ & $0$\\
$g_{\scst S}$ & $v_0$ & & & $v_0v_2-v_1v_2-v_0v_1$&  &  & &\\
$\phi_{\scst S}$ & $0$  & $2\gamma_{(v_0,v_i)}$ & $0$ & $0$ & $2v_4v_6v_8$ & $2\gamma'^{v}_{(z)}$ & $0$ & $0$\\
\end{tabular}
\end{center}

\noindent where 
$\gamma'''_{z}$ is $0$ or $\pm v_0v_2$
if the arrow drawn over $z$ in Fig. \ref{6}.c) is, respectively, red or green.  
$\gamma'^{v}_{(z)}$
is a `path' from $z$ following the arrows drawn in Fig. \ref{6}.c).
\end{example}

The following proposition shows that AT-models over ${\bf Z}/p$, where $p$ is prime, can also
be computed using Alg. \ref{latmodel}.

\begin{prop}
Working with coefficients in ${\bf Z}/p$, where $p$ is prime,
let
$$(C,d_{\scst {\bf Z}/p}),H_{\scst {\bf Z}/p},f_{\scst {\bf Z}/p},g_{\scst {\bf Z}/p},\phi_{\scst {\bf Z}/p},\lambda)$$
be the output of Alg. \ref{latmodel}. Then
$$((C,d_{\scst {\bf Z}/p}),H_{\scst {\bf Z}/p},\lambda^{-1} f_{\scst {\bf Z}/p},g,\lambda^{-1}\phi_{\scst {\bf Z}/p})$$
 is an AT-model over ${\bf Z}/p$.
 Furthermore, $\{g_{\scst {\bf Z}/p}(h): h\in H_{\scst {\bf Z}/p}\}$ is a set of representative cycles of generators of
$H({\cal C};{\bf Z}/p)$.
\end{prop}

In order to obtain a $\lambda$-AT-model with $\lambda$ `small',
the following algorithm can be used as a preprocessing of Alg. \ref{latmodel}. It only considers the case
in which $\langle d(\alpha'), \alpha \rangle = 1$. The output of this algorithm is a chain contraction of the given chain complex
${\cal C}$ to another chain complex ${\cal C'}$ with less number of generators. We can compute
 a $\lambda$-AT-model for ${\cal C'}$ applying Alg. \ref{latmodel} to ${\cal C'}$. We can obtain a
$\lambda$-AT model for ${\cal C}$, composing the chain contraction of ${\cal C}$ to ${\cal C'}$ with the
$\lambda$-AT-model for ${\cal C}'$ obtained before, using Formula (\ref{composition})
 given in page \pageref{lchain}.

\begin{alg}\label{pre}
Preprocessing.

\begin{tabbing}
{\sc Input:} \= {\tt A chain complex ${\cal C}=(C,d)$ of dim $n$.}\\\\
 {\tt $C'_0:=C_0$, $d'_0=d_0$, $A:=C_0$, $f_0:=id_{C_0}$, $g_0:=id_{C'_0}$,
   $\phi_0:=0$.} \\
   {\tt For}\= { \tt $q=1$ to $q=n$ do}\\
\> {\tt $C'_q:=\{ \, \}$, $x:=0$, $\gamma:=0$, $\gamma':=0$.}\\
\> {\tt While}\= { \tt $min\{|\langle f_{q-1}d_q(c), c' \rangle|\neq 0:\;  c\in C_{q},\, c'\in C'_{q-1}
\}=1$ do}\\
\>\> {\tt Take}\={ \tt $\alpha\in C'_{q-1}, \;\beta\in C_{q}$ s.t.
$|\langle f_{q-1}d_q(\beta), \alpha \rangle|=1$ then}\\
\>\>\>{\tt $C'_{q-1}:=C'_{q-1} \setminus \{\alpha\},\, A:=A \cup \{\beta\},\,f_q(\beta):=0,\, \phi_q(\beta):=0,$}\\
\>\>\> {\tt $x:=\langle f_{q-1}d_q(\beta),\alpha \rangle,\;\gamma:=f_{q-1}d_q(\beta), \;
\gamma':=\beta - \phi_{q-1}d_q(\beta)$.}\\
\>\>\> {\tt For} \= { \tt each $b \in C_{q-1}$ do}\\
\>\>\>\>{\tt $f_{q-1}(b):=f_{q-1}(b) -x\langle f_{q-1} (b), \alpha \rangle \gamma$,}\\
\>\>\>\>{\tt $\phi_{q-1}(b):= \phi_{q-1}(b) +x\langle f_{q-1} (b), \alpha \rangle \gamma'$,}\\
\>{\tt For}\= { \tt each $a \in C_q\setminus A$ do}\\
\>\> {\tt $C'_q:=C'_q \cup \{a\},\; A:=A\cup\{a\},$}\\
\>\>{\tt $d'_q(a):=f_{q-1}d_q(a)$,  $f_q(a):= a$, $g_q(a):=a - \phi_{q-1}d_q(a)$, $\phi_q(a):=0$.}\\
\\
\textsc{Output:} \= {\tt The set
$((C,d),(C'.d'),f,g,\phi,\lambda)$.}
\end{tabbing}
\end{alg}

\begin{thm}\label{pre-theorem}
The set $((C,d ),(C',d'),f,g,\phi)$  obtained applying Alg. \ref{pre}
defines a chain contraction of the chain complex ${\cal C}=(C,d)$ to the chain complex ${\cal C'}=(C',d')$.
\end{thm}

The proof of this theorem is given in the appendix.

\section{Extracting Integer Homological Information}
As showed before, a $\lambda$-AT-model for a given chain complex ${\cal C}$ provides  information
of the free subgroup of $H({\cal C};{\bf Z})$ as well as the
 prime numbers $p$ involved in the invariant factors of $H({\cal C};{\bf Z})$.
In order to obtain  representative cycles mod $p$ of generators of the free and the torsion subgroups of
$H({\cal C};{\bf Z})$
 we have to compute AT-models for ${\cal C}$ over  ${\bf Z}/p$, for each prime $p$  dividing $\lambda$.

\begin{alg}\label{final} Computing integer homology information and representative cycles of
homology generators of a chain complex ${\cal C}$.

\begin{tabbing}\label{ultimo}
{\sc Input:} \= {\tt a  chain complex ${\cal C}=(C,d)$ of dim $n$.}\\\\
{\tt Apply} \= {\tt Alg. \ref{latmodel} with coefficients in ${\bf Z}$
for computing a}\\
\> {\tt   $\lambda$-AT-model  $((C,d),H,f,g,\phi,\lambda)$ for ${\cal C}$;}\\
\> {\tt For }\= {\tt $q=0$ to $q=n$ do}\\
 \>\>{\tt $\beta_q:=$ number of elements of $H_q$;}\\
 \>{\tt $G:=\{g(h):\; h\in H\}$.}\\
 {\tt For} \= {\tt each prime $p$ dividing $\lambda$ do}\\
 \> {\tt Apply} \= {\tt Alg. \ref{latmodel} with coefficients in ${\bf Z}/p$
 to compute an }\\
 \>\> {\tt  AT-model for ${\cal C}$ over ${\bf Z}/p$: $((C,d_{\scst {\bf Z}/p}),H_{\scst {\bf Z}/p},
 f_{\scst {\bf Z}/p},g_{\scst {\bf Z}/p},\phi_{\scst {\bf Z}/p})$;}\\
 \>\>{\tt $\beta_{(q,p)}:=$ number of elements of $H_{\scst {\bf Z}/p}$ in dim $q$.}\\
 \>\> {\tt  $T_{(0,p)}=\beta_{(0,p)}-\beta_0$;}\\
 \>\>{\tt For }\= {\tt $q=1$ to $n$ do}\\
 \>\>\> {\tt  $T_{(q,p)}=\beta_{(q,p)}-\beta_q-T_{(q-1,p)}$,}\\
  \>\>\>{\tt $G_{\scst {\bf Z}/p}=\{g_{\scst {\bf Z}/p}(h):\; h\in H_{\scst {\bf Z}/p}\}$.}\\
  {\sc Output:} \= {\tt The sets $G$,
 $\{\beta_1,\dots,\beta_n\}$,}\\
\>{\tt  $\{G_{\scst {\bf Z}/p}:\, p$ is a prime dividing $\lambda\}$,}\\
 \> {\tt  and $\{T_{(q,p)}:\, 0\leq q \leq n$ and
 $p$ is a prime dividing $\lambda\}$.}
   \end{tabbing}
\end{alg}

After computing Alg. \ref{ultimo} for a given chain complex ${\cal C}$, we obtain:
\begin{itemize}
\item the Betti numbers $\beta_q$ for $0\leq q\leq n$, and a set $G$ of
independent non-boundary cycles of ${\cal C}$ over ${\bf Z}$ (in fact, $G$ is also
a set of generators of $H({\cal C}; {\bf Q})$);
\item the prime numbers $p$ involved in the invariant factors corresponding to the torsion subgroup
of $H({\cal C},{\bf Z})$, the amount of invariant factors in each dim $q$ that are a power of $p$, $T_{(q,p)}$,
and a set $G_{\scst {\bf Z}/p}$ of representative cycles of generators of  $H({\cal C};{\bf Z}/p)$,
for each prime $p$  dividing $\lambda$.
\end{itemize}

This algorithm uses Alg. \ref{latmodel} over ${\bf Z}$ for computing a $\lambda$-AT-model for ${\cal C}$ and over ${\bf Z}/p$ for computing AT-models for ${\cal C}$, for each prime $p$ dividing $\lambda$. Since the complexity of Alg. \ref{latmodel} is $O(m^3)$, then the complexity of  Alg. \ref{ultimo} is $O(m^3\psi(\lambda))$ in the worst case,
$\psi$ being  the Euler function.

\begin{example}
In Example \ref{exklein}, we have applied Alg. \ref{latmodel} and
computed a $2$-AT-model for ${\cal C}(S)$. We obtained that $H_{\scst S}=\{v_0,v_0v_2\}$ and then the
Betti numbers of ${\cal C}$ are $\beta_0=1$, $\beta_1=1$ and
$\beta_2=0$.
Now, apply  Alg. \ref{latmodel}
 with coefficients in ${\bf Z}/2$ to obtain an AT-model 
  $((C,d_{\scst{\bf Z}/2})),H_{\scst{\bf Z}/2},f_{\scst{\bf Z}/2},$ $ g_{\scst{\bf Z}/2},\phi_{\scst{\bf Z}/2})$
where $H_{\scst{\bf Z}/2}=\{v_0,v_0v_2,v_0v_4,v_0v_3v_8\}$ for ${\cal C}(S)$. Then, $\beta_{(0,2)}=1$,
$\beta_{(1,2)}=2$, $\beta_{(2,2)}=1$.
Therefore, $T_{(0,2)}=0$, $T_{(1,2)}=1$ and $T_{(2,2)}=0$.

We conclude that
$H_0(S;{\bf Z})\simeq {\bf Z}$ and $H_1(S;{\bf Z})\simeq {\bf Z}\oplus {\bf Z}/2$.
\end{example}

\begin{figure}[t!]
\begin{center}\includegraphics[width=5.2cm]{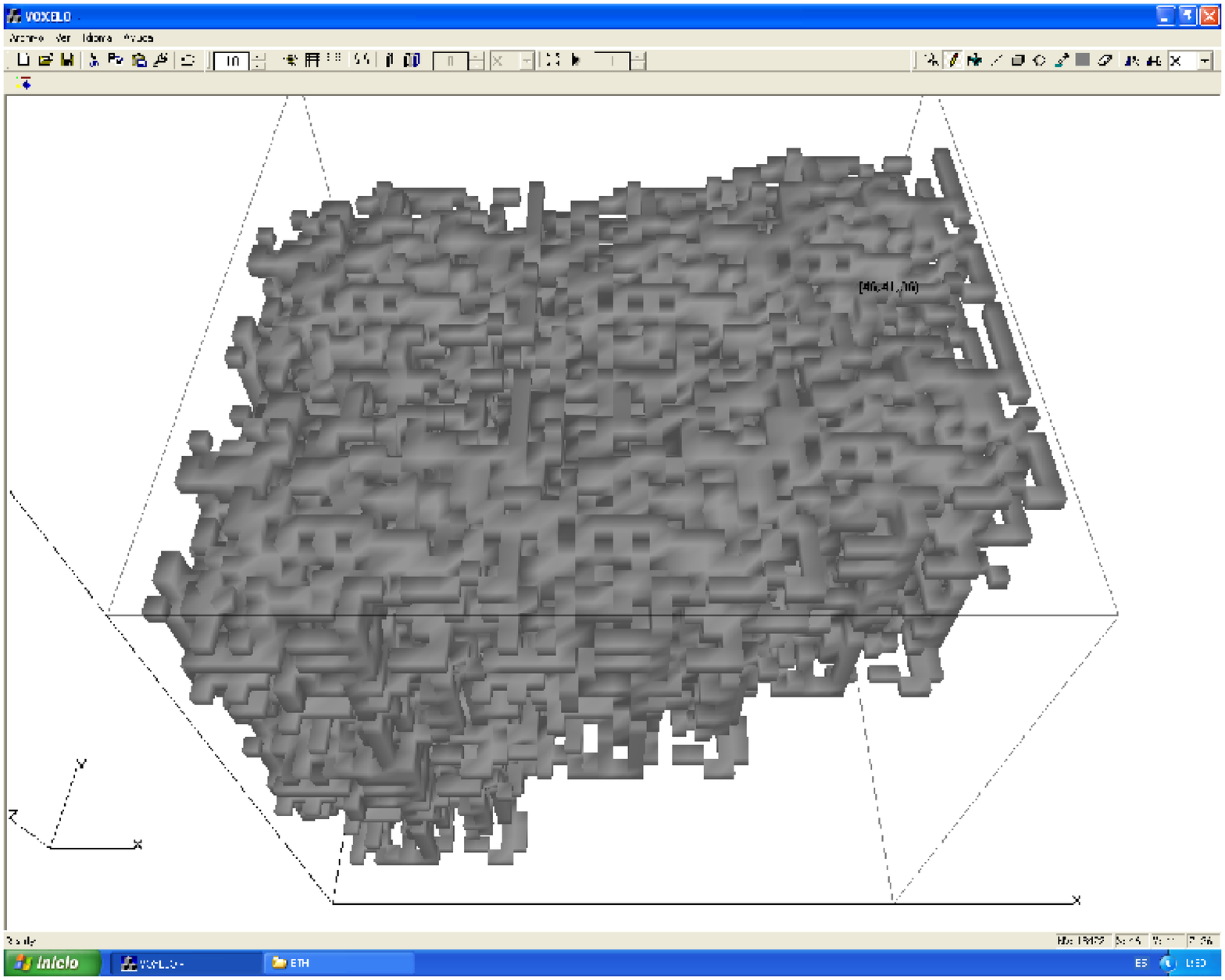}
\includegraphics[width=5.2cm]{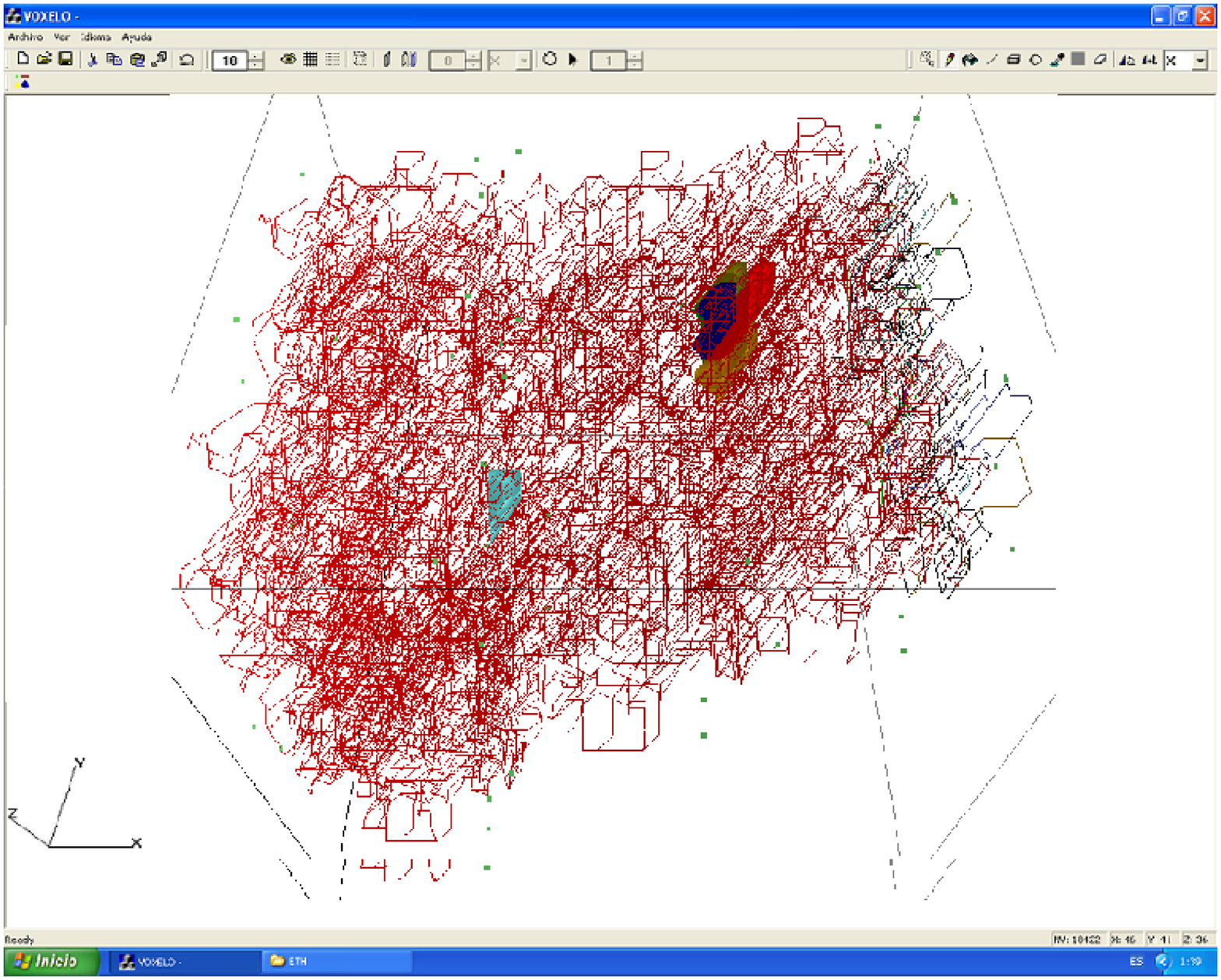}
\end{center}
 \label{7}\caption{A porous 3D digital image and representative cycles of its $2709$ black holes.}
\end{figure}

Alg. \ref{ultimo} has been implemented \cite{javier}. Fig. \ref{7} shows  a 3D digital image with $18491$ black voxels.  In $144$ seconds
we obtain that the image has $51$ black connected components, $2709$ black holes and $8$ black cavities.

\section{Conclusions and Future Work}
We have constructed  an algebraic tool, called
$\lambda$-AT-model, in order to compute topological features of
$n$-dimensional objects (for instance, nD digital pictures).
Associating an algebraic representation to the initial object, a chain complex,  the
algorithm presented here provides
the free subgroup of homology, with integer coefficients, as
well as the primes $p$ that are the candidates to be involved in
the torsion subgroup. In fact, such a set of candidates is the
one of divisors of $\lambda$. We have determined
the minimum $\lambda$ valid for the existence of a
$\lambda$-AT-model.
We  have also presented an algorithm which gives a $\lambda$-AT-model
with `smaller' $\lambda$ than the one given in \cite{GJMR07}.
As for the complexity, the algorithm presented is performed in $O(m^3\psi(\lambda))$ in the worst case,
$\psi$ being  the Euler function and $m$ being the number of cells of a cell decomposition of the object. In order to improve the complexity, 
 first compute a chain contraction from the initial chain complex  using Alg. \ref{pre} to
a `smaller' one with same integer homology and then compute the
$\lambda$-AT-model for the latter.
Afterwards,
 in order to obtain the amount of invariant factors
that are a power of $p$ and representative cycles mod $p$ of
homology generators,
for each $p$  we
compute an AT-model with coefficients in ${\bf Z}/p$.

Future work can be addressed to obtain generators of the torsion
subgroup
of  $H({\cal C};{\bf Z})$, with integer coefficients. Another task is
to study the computation of cohomology features over ${\bf Z}$ from a
$\lambda$-AT-model.

\section*{Appendix: Proof of Theorem \ref{pre-theorem}}

Let $A=\{a^1,\dots,a^m\}$ such that if $i<j$ then $a^i$ was added to $A$ before $a^j$ when the algorithm was applied.  
Fixed $i$,   
assume that 
$f_{r-1}d_r(a^k)=d'_r f_r (a^k)$,
$a^k - g_r f_r (a^k)=\phi_{r-1}d_r(a^k)+
d_{r+1}\phi_r(a^k)$, $f_{r+1} \phi_r(a^k)=0$ and,
if $a^k\in C'_r$, $d_r g_r(a^k)=g_{r-1} d'_r g_r(a^k)$ and
$f_r g_r(a^k)=a^k$, for $1\leq k<i$ and $r=dim (a^k)$.
 
Let $q=dim(a^i)$. We have to prove that the redefined maps that we will denote here by $d''$ (the redefined map of $d'$), 
 $f'$, $g'$ and $\phi'$ satisfy that $f'_{r-1}d_r(a^k)=d''_r f'_r (a^k)$,
$a^k - g'_r f'_r (a^k)=\phi'_{r-1}d_r(a^k)+
d_{r+1}\phi'_r(a^k)$, $f'_{r+1} \phi'_r(a^k)=0$ and,
if $a^k\in C'_r$, $d_r g'_r(a^k)=g'_{r-1} d''_r g'_r (a^k)$ and
$f'_r g'_r(a^k)=a^k$, for $1\leq k<i$ and $r=dim (a^k)$.

We have to consider two cases. (1) If $a^i$ was added to $A$ while 
$min\{|\langle f_{q-1}d_q(c),$ $ c' \rangle|\neq 0:\;  c\in C_{q},\,c'\in C'_{q-1}
\}=1$, then  $d''=d'$ and $g'=g$ (that is, $d'$ and $g$ do not change). 
Therefore, if $a^k\in H_r$ for some $r$, then $d_r g'_r(a^k)=d_r g_r(a^k)=g_{r-1}d'_r(a^k)=g'_{-1}d''_r(a_k)$ for all $1\leq k<i$.
Consider the element $\alpha\in C'_{q-1}$ that defines
the auxiliary parameter $x=\langle f_{q-1}d_q(a^i),\alpha\rangle$ at this stage (observe that $x=\pm 1$, then $xx=1$). 
 
 If $k=i$, then $a^i\not\in C'_q$. We have that $f'_{q-1}d_q(a^i)=f_{q-1} d_q(a^i)-x\langle f_{q-1}d_1(a^i),$ $\alpha\rangle\gamma=
 \gamma-xx\gamma=0=d''_q f'_q(a^i)$  and
  $\phi'_{q-1}d_q(a^i)+d_{q+1} \phi'_q(a^i)=\phi'_{q-1}d_q(a^i)=\phi_{q-1}d_q(a^i)+x\langle f_{q-1}d_q(a^i),\alpha\rangle \gamma'=
 \phi_{q-1}d_q(a^i)+xx(a^i-\phi_{q-1}d_q(a^i))=a^i$. Moreover, 
 $f'_{q+1} \phi'_q(a^i)=0$ since $\phi'_q(a^i)=0$.

Observe that: For any $r$-chain $b$ such that $r< q-1$, $f'_r(b)= f_r(b)$ and $\phi'_r(b)= \phi_r(b)$.  $d_q(\gamma')=d_q( a^i-\phi_{q-1}d_q(a^i))= d_q(a^i) -(d_q(a^i)-g_{q-1}f_{q-1}d_q(a^i)-\phi_{q-1} d_{q-1}d_q(a^i))=g_{q-1}f_{q-1}d_q(a^i)=
g_{q-1}(\gamma)$.  For any $q$-chain $b$, $f_q(b)=0$ and $\phi_q(b)=0$, therefore, $f'_q(b)=0$ and $\phi'_q(b)=0$.

For $1\leq k<i$. (1.a) If $r:=dim(a^k)<q-1$ then 
$f'_{r-1}d_r(a^k)=f_{r-1}d_r(a^k)=d'_{r-1} f_{r}(a^k)=d''_{r-1}f'_r(a^k)$ and
$\phi'_{r-1}d_r(a^k)+d_{r+1}\phi'_r(a^k)=\phi_{r-1}d_r(a^k)+d_{r+1}\phi_r(a^k)$ $=
 a^k-g_r f_r(a^k)=a^k-g_r f'_r(a^k)$.
Moreover, $f'_{r+1}\phi'_r(a^k)=f'_{r+1}\phi_r(a^k)=f_{r+1}\phi_r($ $a^k)-x\langle f_{r+1}\phi_r(a^k),\alpha\rangle\gamma)
=0$ by induction.
 If $a^k\in C'_r$ then $f'_r g_r(a^k)=f_r g_r(a^k)$ $= a^k$.
(1.b) If $dim (a^k)=q-1$, then 
$f'_{q-2}d_{q-1}(a^k)=f_{q-2}d_{q-1}(a^k)=d'_{q-1}f_{q-1}(a^k)$ $=d'_{q-1}f_{q-1}(a^k)-x\langle f_{q-1}(b),\alpha\rangle f_{q-2}d_{q-1}d_{q}(\beta)=
 d'_{q-1}f_{q-1}(a^k)-d'_{q-1}(x\langle f_{q-1}(b),$ $\alpha\rangle f_{q-2}d_{q}(\beta))=d'_{q-1}f'_{q-1}(a^k)=d''_{q-1}f'_{q-1}(a^k)$ and
$\phi'_{q-2}d_{q-1}(a^k)+d_{q}\phi'_{q-1}(a^k)=\phi_{q-2}d_{q-1}(a^k)+d_q\phi_{q-1}(a^k)+x\langle f_{q-1}(a^k),\alpha\rangle d_q(\gamma')
= a^k-g_{q-1}f_{q-1}(a^k)+x\langle f_{q-1}($ $a^k),\alpha\rangle g_{q-1}(\gamma)
=a^k-g_{q-1}f'_{q-1}(a^k)$. 
Moreover, $f'_{q}\phi'_{q-1}(a^k)=
f'_q(\phi_{q-1}(a^k)+x\langle f_{q-1}(a^k),\alpha\rangle \gamma')$. 
Now, $f'_q \phi_{q-1}(a^k)=0$ since for any $q$-chain $b$, $f'_q(b)=0$;
$f'_q(\gamma')=f'_q( a^i-\phi_{q-1}d_q(a^i) )=0$ for the same reason than above. Then, $f'_{q}\phi'_{q-1}(a^k)=0$. 
If $a^k\in C'_{q-1}$ then $a^k\neq \alpha$ and  $f'_{q-1} g_{q-1}(a^k)= f_{q-1} g_{q-1}(a^k)-
x\langle f_{q-1}g_{q-1}(a^k),\alpha\rangle \gamma= a^k-x\langle  a^k,\alpha\rangle \gamma=
a^k$.
(1.c) If $dim (a^k)=q$, then 
$f'_{q-1}d_q(a^k)$ $=f_{q-1}d_q(a^k)-x\langle f_{q-1}d_q(a^k),\alpha\rangle\gamma=
d'_q f_q (a^k)-x\langle d'_q f_q(a^k),\alpha\rangle\gamma=0=d''_q f'_q (a^k)$ and
$\phi'_{q-1}d_q(a^k)+d_{q+1}\phi'_q(a^k)=\phi_{q-1}d_q(a^k)+d_{q+1}\phi_q(a^k)
+x\langle f_{q-1}d_q(a^k),\alpha\rangle\gamma'
= a^k-g_q f_q(a^k)+x\langle d'_q f_{q}(a^k),\alpha\rangle\gamma'
= a^k -g_q f_q(a^k)= a^k-g_q f'_q(a^k)$.
Moreover, $f'_{q+1}\phi'_q(a^k)=0$.
 
(2) If $a^i$ was added to $A$ when  $\{|\langle f_{q-1}d_q(c), c' \rangle|\neq 0:\;  c\in C_{q},\, c'\in C'_{q-1}
\}=\emptyset$ or
$min\{|\langle f_{q-1}d_q(c), c' \rangle|\neq 0:\;  c\in C_{q},\, c'\in C'_{q-1}
\}> 1$
 then,    
 $f'_{q-1}d_q(a^i)=f_{q-1}d_q(a^i)=d''_q(a^i)=d''_q f'_q(a^i)$ and 
 $a^i-g'_q f'_q(a^i)=a^i- g'_q(a^i)=
 a^i-(a^i-\phi_{q-1}d_q(a^i))=\phi_{q-1}d_q(a^i)=\phi'_{q-1}d_q(a^i)
+d_{q+1}\phi'_q(a^i)$. Second, 
$d_q g'_q(a^i)=d_q(a^i)-d_q\phi_{q-1}d_q(a^i)=g_{q-1}f_{q-1}d_q(a^i)+  \phi_{q-2}d_{q-1}d_q(a^i)
=g_{q-1}f_{q-1}d_q(a^i)=g_{q-1}d''_q(a^i)=g'_{q-}d''_q(a^i)$ and 
$f'_q g'_q(a^i)=f'_q(a^i-\phi_{q-1}d_q(a^i))=
a^i-f_q\phi_{q-1}d_q(a^i)= a^i$,
since  $f_{r+1} \phi_r(a^k)=0$  for $1\leq k< i$ and $r=dim(a^k)$, by induction. 
Finally, $f'_{q+1}\phi'_q(a^i)=0$ since $\phi'_q(a^i)=0$.

\newpage

\begin{center} STATEMENT\end{center}

\vspace{1cm}

We present some notes in response to the suggestions made by the reviewers.

Concerning the comments of Reviewer 1, the main remark is that we should add further explanations for the readers who are not specialist in the field. We have followed this philosophy. We could classify the suggestions in the following categories: 
\begin{itemize}
\item Little gaps or changes of notations: we have followed all the reviewer's instructions.
\item  Addition of examples: we have added more examples to illustrate some aspects that the reviewer asked to be clarified in this way.       
\item Further explanations: we have recalled more background concepts and detailed some technical aspects at the points that the referee suggested. In some other cases, we have simply extended the given explanations.
\item Regarding Algorithm 1 (Algorithm 13 in the previous version), it has been modified since it was difficult to follow the technical aspects, as pointed out by the reviewer. The changes have allowed to carry out the proof of Theorem 16 (Theorem 14 in the previous version)in a clearer way. 
\item Proof of Theorem 19 (Theorem 18 in the previous version) has been provided in an appendix.
\item With respect to the reviewer's question about the reduction of the complexity using Alg. 2 (Alg. 17 in the previous version), we have to say that it is difficult to quantify such as a reduction in general. We plan to study this question in detail  and  expect to provide experiments and concrete examples of such a reduction. 
 
\end{itemize}
As for the comments of Reviewer 2, all the suggestions have been carried out. Most of them are Grammar errors or little gaps that have been corrected. Also the proposed change in the numbering for definitions, theorems, etc., has been undertaken.

\end{document}